# pyLEMMINGS: Large Margin Multiple Instance Classification and Ranking for Bioinformatics Applications

Amina Asif[1], Wajid Arshad Abbasi[1], Farzeen Munir[2], Asa Ben-Hur[3], Fayyaz ul Amir Afsar Minhas[1,*]

[1]Biomedical Informatics Research Laboratory (BIRL), Department of Computer and Information Sciences, Pakistan Institute of Engineering and Applied Sciences (PIEAS), Islamabad, Pakistan.

[2]Machine Learning and Vision Laboratory, School of Electrical Engineering and Computer Science, Gwangju Institute of Science and Technology, Gwangju, South Korea

[3]Department of Computer Science, Colorado State University, Fort Collins, Colorado, USA.

*To whom correspondence should be addressed.

**Abstract:**

**Motivation:** A major challenge in the development of machine learning based methods in computational biology is that data may not be accurately labeled due to the time and resources required for experimentally annotating properties of proteins and DNA sequences. Standard supervised learning algorithms assume accurate instance-level labeling of training data. Multiple instance learning is a paradigm for handling such labeling ambiguities. However, the widely used large-margin classification methods for multiple instance learning are heuristic in nature with high computational requirements. In this paper, we present stochastic sub-gradient optimization large margin algorithms for multiple instance classification and ranking, and provide them in a software suite called pyLEMMINGS.

**Results:** We have tested pyLEMMINGS on a number of bioinformatics problems as well as benchmark datasets. pyLEMMINGS has successfully been able to identify functionally important segments of proteins: binding sites in Calmodulin binding proteins, prion forming regions, and amyloid cores. pyLEMMINGS achieves state-of-the-art performance in all these tasks, demonstrating the value of multiple instance learning. Furthermore, our method has shown more than 100-fold improvement in terms of running time as compared to heuristic solutions with improved accuracy over benchmark datasets.

**Availability and Implementation:** pyLEMMINGS python package is available for download at: http://faculty.pieas.edu.pk/fayyaz/software.html#pylemmings.

**Contact:** Correspondence should be addressed to Dr. Fayyaz Minhas, fayyazafsar@gmail.com.

**Supplementary Information:** No supplementary files.

## 1 Introduction

Over the past two decades, the use of machine learning in bioinformatics has grown significantly, including tasks such as drug design, protein interaction prediction, binding site prediction etc. (Zhang and Rajapakse 2009; Qi, Bar-Joseph, and Klein-Seetharaman 2006; Gertrudes et al. 2012). Effective predictive models for these problems can help prioritize and optimize the design of wet-lab experiments and reduce their personnel and monetary costs. In conventional supervised machine learning, a predictive model is constructed using a training dataset consisting of labeled examples (Kotsiantis, Zaharakis, and Pintelas

2007). Many machine learning algorithms such as Neural Networks (Anthony and Bartlett 2009), Random Forests (Breiman 2001), Support Vector Machines (Cortes and Vapnik 1995), etc., have proven to be effective in situations where the datasets are sufficiently large and accurately labeled. The requirement that training data be accurately labeled can prove to be unrealistic in many bioinformatics problems for a variety of reasons. Take the discovery of binding sites using deletion mutagenesis for example: this technique discovers regions that are important for binding, but provides no guarantee that the resulting region is the actual binding site. In such cases, it might be easier to provide labels at a higher level of abstraction instead of labeling individual instances. For example, for the task of protein interaction-site prediction, the binding site annotations of the training data may cover an area which is much larger than the minimal set of residues responsible for the interaction (Minhas and Ben-Hur 2012). Furthermore, not all residues in an interaction site may contribute equally to protein binding. This gives rise to ambiguities in the labels of training data.

Multiple instance Learning (MIL) (Dietterich, Lathrop, and Lozano-Pérez 1997) is a supervised machine learning paradigm for handling such labeling ambiguities in training data. In conventional supervised learning, every training example has an associated label. In contrast, MIL training datasets have labels associated to groups of examples called *bags*. A bag is given a positive label if at least one of example in it is positive. However, it is not known which of the examples in the bag are positive. A negative label is assigned to a bag in which all examples are negative. The concept of bags is illustrated in Figure 1. The goal of Multiple Instance Classification is to classify unseen bags or instances based on training data consisting of labeled bags. MIL can be viewed as a generalization of the traditional supervised learning (Kotsiantis, Zaharakis, and Pintelas 2007) as it reduces to conventional learning if each bag contains only one example.

Multiple instance learning was first motivated by the problem of predicting the binding of drug molecules (Dietterich, Lathrop, and Lozano-Pérez 1997). The task was to predict whether a given drug molecule binds strongly to a target protein or not. A drug molecule can exist in multiple conformations and its binding is caused by one or more of its conformations. It is very difficult to experimentally ascertain the exact conformations of a drug molecule that are responsible for its binding. Consequently, binding or non-binding labels can only be assigned to drug molecules and not to their individual conformations. Therefore, this classification problem is not directly solvable using conventional classification techniques. It was, therefore, presented as a multiple instance learning problem with a bag representing a drug molecule. The feature representation of a possible conformation of the molecule is taken as an example in the bag. A positive bag means that at least one of the conformations of the molecule binds the protein whereas a negatively labeled bag implies that none of the conformations of that molecule bind. Dietterich et al. (Dietterich, Lathrop, and Lozano-Pérez 1997) developed a specialized MIL classifier for solving this problem. Nowadays, MIL finds its application in a variety of fields including bioinformatics, computer vision, text and image processing, etc. (Andrews, Tsochantaridis, and Hofmann 2003; Babenko, Yang, and Belongie 2009; Maron and Lozano-Pérez 1998; Hong et al. 2014; Wu et al. 2015).

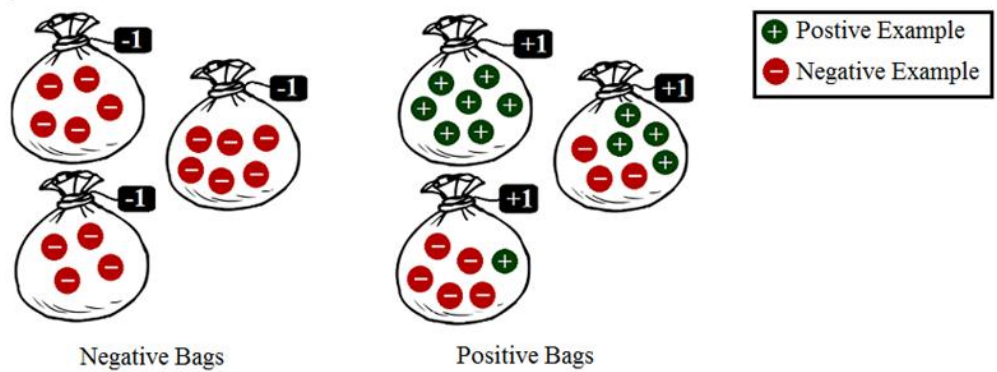

**Figure 1-** Illustration of the concept of Bags. A bag is given a positive label if one or more of its examples are positive. All the examples in a negative bag are negative.

MIL problems require specialized learning algorithms that take training data in the form of labeled bags instead of individual instances. Many MIL methods using different classification approaches have been proposed. These include Learning Axis Parallel Concepts (Dietterich, Lathrop, and Lozano-Pérez 1997), Diverse Density and its Expectation Maximization version (Yang 2005), Boosting (Viola et al. 2005), Citation kNN [16], mi/MI-SVM (Andrews, Tsochantaridis, and Hofmann 2003), MILES (Chen, Bi, and Wang 2006), mi-DS (Nguyen et al. 2013), Generalized Dictionaries for Multiple Instance Learning (Shrivastava et al. 2015), etc. Our focus in this paper is on large margin methods as these offer better generalization (Zhou 2014), especially when the number of features is large and the number of examples is small as is often the case in bioinformatics. MI-SVM and mi-SVM use a large margin formulation of the MIL problem which leads to hard optimization problems which are difficult to solve even for moderately-sized datasets. Andrews et al. have proposed local optimization heuristics for these optimization problems (Andrews, Tsochantaridis, and Hofmann 2003). However, these heuristics do not present a direct solution to the problem and are, consequently, unable to find the optimal solution to the MIL problem in many cases. Furthermore, both mi-SVM and MI-SVM require iterative retraining of a conventional SVM which makes the algorithms very expensive in terms of running times for large datasets. Another method, MILES (Chen, Bi, and Wang 2006), solves the MIL problem by transforming the problem into a standard supervised learning problem. The method does so by first mapping bags into a feature space characterized by instances. The mapping is based on an instance similarity measure. A 1-norm SVM is then used for feature reduction and classification. In comparison to other methods, MILES is more efficient in terms of running times than other methods but its classification performance over benchmark datasets is much lower than the state of the art. Other methods proposed for MIL (Babenko, Yang, and Belongie 2009; Hong et al. 2014; Viola et al. 2005; Wu et al. 2014; Kraus, Ba, and Frey 2016; Huang, Gao, and Zhou 2014) were designed for computer vision and image processing applications and the results over benchmark datasets from bioinformatics domains are not reported.

In this paper, we present a large margin formulation for multiple instance classification and ranking. Due to its large margin properties, our method has tunable complexity via adaptive and controllable Vapnik Chervonenkis Dimensions and better generalization performance (Bartlett and Shawe-Taylor 1999) in comparison to other approaches. We propose a stochastic sub-gradient descent based solver for large margin MIL. Our method is inspired from PEGASOS (Shalev-Shwartz, Singer, and Srebro 2007) which is a stochastic sub gradient based solver for conventional SVM. For SVM based methods, non-linearity of the classifier is usually achieved through kernelization (Hofmann, Schölkopf, and Smola 2008). Using conventional kernelization in PEGASOS yields a running time that is dependent on the size of the dataset and does not remain feasible when the dataset is very large. As an alternative, we have proposed a novel locally linear encoding (Ladicky and Torr 2011) based formulation and its solution using stochastic sub-gradient optimization. The proposed algorithm and its performance on benchmark datasets is presented in the forthcoming sections. We have developed an open-source implementation of the proposed algorithms called pyLEMMINGS. Details of the software are presented in section 2.4. The software is available online at the URL: http://faculty.pieas.edu.pk/fayyaz/software.html#pylemmings. Apart from testing and evaluating the algorithms over benchmark datasets, we have used pyLEMMINGS over three practical bioinformatics problems: Localization of the binding site of Calmodulin binding proteins, Identification of prion forming domains and classification of protein amyloidogenic regions. Our experiments show that modeling these problems through pyLEMMINGS based multiple instance learning gives state-of the art results.

## 2 Methods

In this section, we present mathematical formulations for large margin multiple instance classification and ranking and their solution using Stochastic Sub-gradient Optimization (SSGO) (Shalev-Shwartz, Singer, and Srebro 2007).

## 2.1 Problem Formulation

A typical multiple instance learning problem can be represented as follows: we are given $n$ examples $x_1, x_2, \ldots, x_n$ grouped into $N$ non-overlapping bags $B_1, B_2, \ldots, B_N$ such that each bag $B_I$ contains one or more instances and has an associated label $Y_I$. It is important to note that, in contrast to conventional supervised learning in which each instance is associated with a label, in multiple instance learning, the labels are at the bag level. The objective of MIL is to learn a function $f(B; w)$, parametrized by $w$, that produces a decision score for a given bag $B$. The function must satisfy a set of constraints based on the training data and should generalize to previously unseen test cases. In multiple instance classification, the constraints require that there should be at least one positive example in a positive bag and all examples in a negative bag should be classified as negative. In other words, the highest scoring example in a positive bag should produce a positive classification. All examples in a negative bag should produce negative scores. This can be achieved with a linear decision function that produces a bag level score based on the highest scoring example in the bag. The function can be expressed as $f(B; w) = \max_{x \in B} \langle w, x \rangle$, where $\langle w, x \rangle$ indicates dot-product. For MIL, the classification constraint can then be written as $Yf(B; w) > 0$.

In multiple instance ranking, a bag $B$ has an associated integer rank $Y$. The problem of ranking assumes that there exists a grading trend in the labels of training data, i.e., the decision score produced by bag $B_I$ should be greater than that produced by $B_J$ if rank of $B_I$ is higher than that of $B_J$ i.e., $Y_I > Y_J$. The particular example or examples in a bag that are responsible for its rank are not known. For a pair of bags $B_I$ and $B_J$ such that $Y_I > Y_J$, the constraint for ranking can be written as $f(B_I; w) > f(B_J; w)$.

## 2.2 Linear Multiple Instance Classification and Ranking

An effective machine learning model is one that minimizes both structural risk and empirical errors (Suykens and Vandewalle 1999). A conventional support vector machines achieves this by maximizing the margin and minimizing the loss over training examples. We present a similar formulation for multiple instance classification. The objective function can be mathematically given as,

$$\min_w \rho(B, Y; w) = \frac{\lambda}{2} \|w\|^2 + \frac{1}{N} \sum_I^N l(B_I, Y_I; w)$$

The objective function contains two terms: $\|w\|^2$ is the inverse of the margin and it controls the regularization of the model. The extent of regularization is controlled by the regularization hyperparameter $\lambda > 0$. The other term refers to the overall loss due to misclassifications or margin violations. Here, $l(B_I, Y_I; w)$ is the loss incurred due to classification of bag $B_I$. It should be positive when the bag is misclassified and zero otherwise. The ideal zero-one loss function can be given as $l(B_I, Y_I; w) = \mathbb{I}[Y_I f(B_I; w) < 0]$. Here, $\mathbb{I}[\cdot]$ is the indicator function with $\mathbb{I}[\cdot] = 1$ if the argument is true and 0 otherwise. However, the ideal loss function is non-convex and hence not suitable for gradient-based optimization. Therefore, akin to conventional SVMs, we use hinge loss as a convex approximation of the ideal loss function given by $l(B_I, Y_I; w) = \max\{0, 1 - Y_I f(B_I; w)\} = \max\{0, 1 - Y_I \max_{x \in B_I} \langle w, x \rangle\}$.

To solve the optimization problem, we use stochastic sub-gradient optimization method inspired from the PEGASOS solver for conventional SVMs (Shalev-Shwartz, Singer, and Srebro 2007). In each iteration of the algorithm, a bag $B_I$ is selected at random from the training set. The most positive example in the bag is determined as $x_I = \mathrm{argmax}_{x \in B_I} \langle w, x \rangle$. The sub-gradient of the objective function $\rho(B, Y; w)$ is calculated for the chosen bag with respect to the weight vector $w$. The weight vector is then updated in the direction opposite to the direction of the sub-gradient. The complete derivation is described below. Suppose a bag $B_{I_t}$ is chosen at iteration $t = 1 \ldots T$, then the objective function with respect to the bag is given as:

$$\rho(B_{I_t}, Y_{I_t}; w) = \frac{\lambda}{2} \|w_t\|^2 + \max\{0, 1 - Y_{I_t} f(B_{I_t}; w)\}$$

The sub-gradient of the objective function at iteration $t$ is given by

$$\nabla_t = \lambda w_t - \mathbb{I}[Y_{I_t} \langle w_t, x_{I_t} \rangle < 1] Y_{I_t} x_{I_t}$$

This leads to the following weight-update at iteration $t$:
$$w_{t+1} = w_t - \eta_t \nabla_t$$
Here, $\eta_t = \frac{1}{\lambda t}$ is the learning rate that decreases over the iterations. The complete weight update equation thus becomes:
$$w_{t+1} = w_t - \eta_t \left[\lambda w_t - \mathbb{1}[Y_{I_t}\langle w_t, x_{I_t}\rangle < 1]Y_{I_t} x_{I_t}\right]$$

At the $t^{th}$ iteration, the weights are updated according to the above equation. The updated weights are then used in the objective function for the next iteration. The complete algorithm is presented in Figure 2(a).

Linear multiple instance ranking can also be implemented by a simple change of the loss function(Hu, Li, and Yu 2008; Bergeron et al. 2008). In ranking, a positive loss should be incurred if the decision function scores of two bags do not correspond to their ranks, i.e., for bags $I$ and $J$ such that $Y_I > Y_J$, the loss function should produce a positive loss if $f(B_I; w) < f(B_J; w)$. Also, the loss should be proportional to the difference in the actual ranks of the two bags $(Y_I - Y_J)$. For a pair of bags $B_I, B_J$ with $Y_I > Y_J$, such a convex loss function can be written as $l(B_I, B_J; w) = \max\{0, (1 - f(B_I; w) + f(B_J; w))\}(Y_I - Y_J)$. The loss will be positive when the ranking constraint is violated within a margin, i.e., $f(B_I; w) < 1 + f(B_J; w)$. If there are $M$ pairs of bags, the objective function can be written as

$$\min_w \rho(B, Y; w) = \frac{\lambda}{2}\|w\|^2 + \frac{1}{M}\sum_{I,J: Y_I > Y_J}^{M} \max\{0, (1 - f(B_I; w) + f(B_J; w))\}(Y_I - Y_J)$$

The optimization problem is solved using the same approach as mentioned earlier for classification. The difference here is that, instead of choosing only one bag stochastically for updating the weights, we now choose a pair of bags $B_I, B_J$ such that $Y_I > Y_J$. The final weight update equation thus becomes,

$$w_{t+1} = w_t - \eta_t \left[\lambda w_t - \mathbb{1}[f(B_{I_t}; w_t) < 1 + f(B_{J_t}; w_t)]\left(x_{J_t} - x_{I_t}\right)(Y_{I_t} - Y_{J_t})\right]$$

The complete algorithm is given in Figure 2(b).

| (a) | (b) |
|---|---|
| **INPUT:** Bags $B_1, B_2, \ldots, B_N$ with their labels $Y_1, Y_2, \ldots, Y_N$, $\lambda$: the regularization parameter and $T$: the number of iterations | **INPUT:** Bags $B_1, B_2, \ldots, B_N$ with their associated ranks $Y_1, Y_2, \ldots, Y_N$, $\lambda$: the regularization parameter and $T$: the number of iterations |
| **INITIALIZE:** $w_1 = 0$ | **INITIALIZE:** Set $w_1 = 0$ |
| For t = 1,2,3, …, T: | For t = 1,2,3, …, T: |
|    Choose a bag $B_t$ uniformly at random |    Choose a bag pair $B_{I_t}, B_{J_t}$ such that $Y_{I_t} > Y_{J_t}$ |
|    $x_t^* = $ Highest scoring example from $B_t$ |    $x_{I_t}^* = $ Highest scoring example in $B_{I_t}$ |
|    $\eta_t = \frac{1}{\lambda t}$ |    $x_{J_t}^* = $ Highest scoring example in $B_{J_t}$ |
|    If $Y_t\langle w_t, x_t^*\rangle < 1$: |    $\eta_t = \frac{1}{\lambda t}$ |
|       $w_{t+1} = (1 - \eta_t \lambda)w_t + \eta_t Y_t x_t^*$ |    If $\langle w_t, x_{I_t}^*\rangle - \langle w_t, x_{J_t}^*\rangle < 1$: |
|    Else: |       $w_{t+1} = (1 - \eta_t \lambda)w_t$ |
|       $w_{t+1} = (1 - \eta_t \lambda)w_t$ |       $\quad + \eta_t \left(x_{J_t}^* - x_{I_t}^*\right)(Y_{I_t} - Y_{J_t})$ |
| **OUTPUT:** $w_{T+1}$ |    Else: |
| |       $w_{t+1} = (1 - \eta_t \lambda)w_t$ |
| | **OUTPUT:** $w_{T+1}$ |

**Figure 2-** Stochastic Sub-gradient Optimization based algorithms for Large Margin (a) Binary Classification and (b) Ranking.

## 2.3 Locally Linear Multiple Instance Classification and Ranking

Conventionally, nonlinearity in the classification boundaries of large margin methods is achieved through kernelization. However, this may become infeasible for large data sets. Locally Linear Support Vector Machines (Ladicky and Torr 2011) presents an alternative approach to non-linear classification using locally linear functions through locally linear coding (Yu, Zhang, and Gong 2009). Based on the same concept, we propose locally linear coding based multiple instance learning. We first provide a brief description of locally linear coding followed by the formal representation of locally linear multiple instance classification and ranking.

In locally linear coding, a point in a data manifold can be represented as a linear combination of some *anchor points* (Yu, Zhang, and Gong 2009). At a conceptual level, anchor points can be considered as sampling points on the surface of the manifold spanned by a given data set and they can be chosen through random sampling or clustering. For a $d$-dimensional feature space, anchor points can be represented by a $d \times K$ matrix $\boldsymbol{V} = [\boldsymbol{v_1}\ \boldsymbol{v_2}\ ...\ \boldsymbol{v_K}]$. A point $\boldsymbol{x}$ can now be written as the linear combination $\boldsymbol{x} \approx \boldsymbol{V}\boldsymbol{\gamma}(\boldsymbol{x})$. Here, $\boldsymbol{\gamma}(\boldsymbol{x}) = [\gamma_1(\boldsymbol{x})\ \gamma_2(\boldsymbol{x})\ ...\ \gamma_K(\boldsymbol{x})]^T$ is the vector of $K$ local coordinates of point $\boldsymbol{x}$. In locally linear coding, local coordinates are obtained by minimizing the reprojection or distortion error $\|\boldsymbol{x_i} - \boldsymbol{V}\boldsymbol{\gamma}(\boldsymbol{x_i})\|^2$ and enforcing local sparsity which is achieved by forcing small values for local coordinates corresponding to far-away anchor points. This is achieved by solving the following optimization problem: $min_\gamma \sum_i \|\boldsymbol{x_i} - \boldsymbol{V}\boldsymbol{\gamma}(\boldsymbol{x_i})\|^2 + \sigma \sum_{k=1}^{K} \gamma_k(\boldsymbol{x_i})^2 \|\boldsymbol{v_k} - \boldsymbol{x_i}\|^2$ (Ladicky and Torr 2011). Here, $\sigma > 0$ is a hyperparameter that controls the tradeoff between reconstruction error and sparsity. A closed-form solution of this optimization problem makes calculation of local coordinates very easy and computationally efficient.

One of the most important properties of locally coding is that any smooth function $w(\boldsymbol{x})$ can be approximated by a linear combination of the function values for the anchor points, i.e., $w(\boldsymbol{x}) \approx \sum_{k=1}^{K} w(\boldsymbol{v_k})\gamma_k(\boldsymbol{x})$ or $w(\boldsymbol{x}) \approx w(\boldsymbol{V})^T \boldsymbol{\gamma}(\boldsymbol{x})$ where $w(\boldsymbol{V}) = [w(\boldsymbol{v_1})\ \ w(\boldsymbol{v_1})\ \ ...\ \ w(\boldsymbol{v_K})]^T$. As in locally linear SVM (Ladicky and Torr 2011) and context-aware feature mapping (Minhas, Asif, and Arif 2016), we utilize this property of local coding to obtain a locally linear approximation of the decision function. This is achieved by making the weight vector local or context dependent, i.e., instead of having a fixed weight vector in the decision function $\langle \boldsymbol{w}, \boldsymbol{x} \rangle$, we use a local weight vector function $\boldsymbol{w}(\boldsymbol{x})$ so that the decision function $\langle \boldsymbol{w}(\boldsymbol{x}), \boldsymbol{x} \rangle$ is locally linear. Mathematically, the local weight vector $\boldsymbol{w}(\boldsymbol{x})$ can be written as a linear combination of the weight values evaluated over anchor points, i.e., $\boldsymbol{w}(\boldsymbol{x}) = \sum_{k=1}^{K} \boldsymbol{w}(\boldsymbol{v_k})\gamma_k(\boldsymbol{x})$ or $\boldsymbol{w}(\boldsymbol{x}) = \boldsymbol{W}\boldsymbol{\gamma}(\boldsymbol{x})$ where, $\boldsymbol{W}$ is the $d \times K$ matrix containing local weight values for the anchor points $\boldsymbol{W} = [\boldsymbol{w}(\boldsymbol{v_1})\ \boldsymbol{w}(\boldsymbol{v_2})\ ...\ \boldsymbol{w}(\boldsymbol{v_K})]$.

We now describe the use of the local weight function for locally linear multiple instance learning. Given a dataset, we first obtain the local coordinates of all examples using a fixed and a prior chosen number $K$ of anchor points. Following the case for linear multiple instance ranking, the locally linear discriminant function for a bag $\boldsymbol{B_I}$ can now be written as: $f(\boldsymbol{B}; \boldsymbol{w}) = \max_{x \in B} \langle \boldsymbol{w}(\boldsymbol{x}),\ \boldsymbol{x} \rangle = \max_{x \in B} \langle \boldsymbol{W}\boldsymbol{\gamma}(\boldsymbol{x}),\ \boldsymbol{x} \rangle$. The objective of multiple instance learning now is to find the weight matrix $\boldsymbol{W}$ by minimizing the following objective function:

$$\min_{\boldsymbol{W}} \rho(\boldsymbol{B}, \boldsymbol{Y};\ \boldsymbol{W}) = \frac{\lambda}{2} \|\boldsymbol{W}\|^2 + \frac{1}{N} \sum_{I}^{N} l(\boldsymbol{B_I}, Y_I;\ \boldsymbol{W})$$

Similar to the linear case, we use hinge loss function $l(\boldsymbol{B_I}, Y_I; \boldsymbol{W}) = \max\{0, 1 - Y_I f(\boldsymbol{B_I}; \boldsymbol{W})\}$ in the above formulation. This problem can also be solved using iterative stochastic sub-gradient optimization as discussed in section 2.2 to yield the following weight update equation:

$$\boldsymbol{W}_{t+1} = \boldsymbol{W}_t - \eta_t \left[ \lambda \boldsymbol{W}_t - \mathbb{I}\left[Y_{I_t} f(\boldsymbol{B}_{I_t}; \boldsymbol{W}) < 1\right] Y_{I_t} \boldsymbol{x}_{I_t} \boldsymbol{\gamma}(\boldsymbol{x}_{I_t})^T \right]$$

Here, $x_{I_t}$ is the highest scoring example from the randomly chosen bag $B_{I_t}$ i.e., $x_{I_t} = \text{argmax}_{x_{I_t} \in B_{I_t}} f(B_{I_t}; W)$.

The objective function for locally linear ranking can be written in a similar fashion. Specifically, given pairs of bags $B_I, B_J$ such that $Y_I > Y_J$, the optimization problem for locally linear ranking becomes:

$$\min_W \rho(B, Y; W) = \frac{\lambda}{2}\|W\|^2 + \frac{1}{M} \sum_{I,J: Y_I > Y_J}^{M} l(B_I, B_J; W)$$

The loss function $l(B_I, B_J; W) = \max\{0, 1 - f(B_I; W) + f(B_J; W)\}(Y_I - Y_J)$ is used. It produces a positive loss proportional to the difference in original ranks when the score produced for $B_J$ is higher than that for $B_I$. Solving this problem using stochastic sub-gradient optimization, we get the following weight update equation:

$$W_{t+1} = W_t - \eta_t \left[\lambda W_t - \mathbb{I}[f(B_{I_t}; W) < 1 + f(B_{J_t}; W)]\left(x_{J_t}\gamma(x_{J_t})^T - x_{I_t}\gamma(x_{I_t})^T\right)(Y_{I_t} - Y_{J_t})\right]$$

$x_{I_t}$ and $x_{J_t}$ are the highest scoring examples from randomly chosen bags $B_{I_t}$ and $B_{J_t}$, respectively. The complete algorithms are presented in Figure 3.

**(a)**
**INPUT:** Bags $B_1, B_2, \ldots, B_N$ with their labels $Y_1, Y_2, \ldots, Y_N$, $\lambda$: regularization parameter $T$: number of iterations and $V$: set of anchor points
**INITIALIZE:** $W_1 = 0$
Compute the local coordinates $\gamma_x$ for each example $x$ based on $V$
For $t = 1,2,3, \ldots, T$:
  Choose a bag $B_t$ uniformly at random
  $x_t^* = $ Highest scoring example from $B_t$
  $\eta_t = \frac{1}{\lambda t}$
  If $Y_t \langle W\gamma_{x_t^*}, x_t^* \rangle < 1$:
    $W_{t+1} = (1 - \eta_t \lambda)W_t + \eta_t Y_t x_t^* \gamma_{x_t^*}^T$
  Else:
    $W_{t+1} = (1 - \eta_t \lambda)W_t$
**OUTPUT:** $W_{T+1}$

**(b)**
**INPUT:** Bags $B_1, B_2, \ldots, B_N$ with their associated ranks $Y_1, Y_2, \ldots, Y_N$, $\lambda$: regularization parameter $T$: number of iterations and $V$: set of anchor points
**INITIALIZE:** $W_1 = 0$
Compute the local coordinates $\gamma_x$ for each example $x$ based on $V$
For $t = 1,2,3, \ldots, T$:
  Choose a bag pair $B_{I_t}, B_{J_t}$ such that $Y_{I_t} > Y_{J_t}$
  $x_{I_t}^* = $ Highest scoring example in $B_{I_t}$
  $x_{J_t}^* = $ Highest scoring example in $B_{J_t}$
  $\eta_t = \frac{1}{\lambda t}$
  If $\langle W\gamma_{x_{I_t}^*}, x_{I_t}^* \rangle - \langle W\gamma_{x_{J_t}^*}, x_{J_t}^* \rangle < 1$:
    $W_{t+1} = (1 - \eta_t \lambda)W_t + \eta_t \left(x_{J_t}\gamma_{x_{J_t}^*}^T - x_{I_t}\gamma_{x_{I_t}^*}^T\right)(Y_{I_t} - Y_{J_t})$
  Else:
    $W_{t+1} = (1 - \eta_t \lambda)W_t$
**OUTPUT:** $W_{T+1}$

**Figure 3-** Stochastic Sub-gradient Optimization based algorithms for Locally Linear (a) Binary Classification and (b) Ranking

## 2.4 Implementation

We have developed an open source software package, pyLEMMINGS (Python LargE Margin Multiple INstance learninG System), in python that implements the large margin methods for MIL discussed in the previous section. pyLEMMINGS can be used for linear and locally linear classification and ranking. Apart from an object-oriented implementation of the proposed learning methods, the package provides built-in modules for k-fold and leave one out cross-validation. pyLEMMINGS supports sparse data as well as parallelization for cross-validation and training. PyLEMMINGS is available for download at

http://faculty.pieas.edu.pk/fayyaz/software.html#pylemmings. The usage instructions have been made available with the package along with example code.

## 3   Experimental Setup

For comparison with other MIL algorithms, we have evaluated our algorithms over the MUSK datasets. MUSK-1 and 2 are widely-used benchmark datasets for the evaluation of MIL algorithms. The datasets were first modeled using multiple instance learning in (Dietterich, Lathrop, and Lozano-Pérez 1997). The datasets are available online at the University of California, Irvine (UCI) repository of machine learning datasets. Both datasets consist of a set of molecules which have been labeled positive or negative by domain experts. A positive label over a molecule indicates that the molecule has a *musky* nature. A molecule may exist in several conformations and it is not known which particular conformation is responsible for the *muskiness* of a positively labeled molecule. A conformation is considered an individual example. All the conformations of a molecule are grouped into a bag and a label is assigned to the bag. Details of the two datasets are presented in Table 1.

We have used 10 fold cross-validation (Refaeilzadeh, Tang, and Liu 2009) with classification accuracy and Area Under Receiver Operating Characteristic Curve (AUC-ROC) (Fawcett 2006) as the performance metrics for evaluation. Average accuracy with standard deviation over 5 runs is reported.

Performance of the algorithms with respect to running times has also been evaluated. Running times in seconds for 10-fold cross-validation runs have been recorded on a i5 personal computer with 2.20 GHz processor for pyLEMMINGS algorithms and mi/MI-SVM algorithms. The implementation for mi-SVM and MI-SVM has been taken from the URL: https://github.com/garydoranjr/misvm (Doran 2017). For all learning tasks, the hyper-parameters are tuned using k-fold cross-validation.

**Table 1**- Details of the benchmark datasets

| Dataset | Instances | Bags | Positive Bags | Negative Bags | Dimensions |
|---|---|---|---|---|---|
| **MUSK-1** | 476 | 92 | 47 | 45 | 166 |
| **MUSK-2** | 6598 | 102 | 39 | 63 | 166 |

## 4   Case Studies

We have used pyLEMMINGS in three multiple instance learning problems from the domain of Bioinformatics. Here, we present a brief description of these problems and the performance of pyLEMMINGS for their solution. The three problems over which the software has been tested are:
  i.   Binding Site Prediction in Calmodulin (CaM) binding proteins (Abbasi et al. 2017)
  ii.  Prediction of prion-forming proteins (Minhas, Ross, and Ben-Hur 2017)
  iii. Prediction of Amyloid Cores

   As shown in Figure 4, in all these problems, we are interested in predicting a local region of a protein sequence that is responsible for a certain property or function of the protein (CaM binding, prion formation or amyloidogenesis). For each problem, we have a training set that contains annotations of protein sequences that indicate these regions of interest based on wet-lab experiments. However, due to the limitations of wet-lab assays, these annotations are not precise and typically cover an area larger than the minimal set of residues responsible for protein function (Minhas and Ben-Hur 2012). Furthermore, not all amino acids in the annotated region contribute equally to the function of the protein (Ng and Henikoff 2003). A naïve method for making such a predictor is to take fixed-length sequence windows within annotated regions as positive examples and the remaining ones as negative and use a conventional classifier such as an SVM or random forests for classification. However, this approach fails to capture labeling ambiguities in the data as not all positively labeled windows are truly positive. Consequently, these problems are ideal for multiple instance learning.

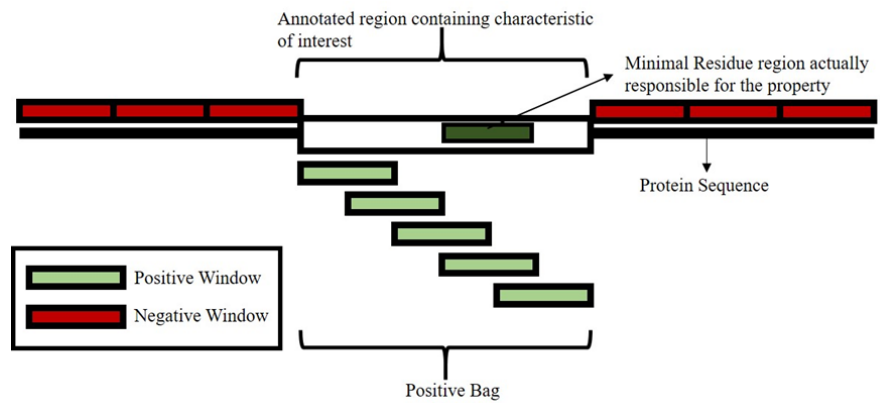

**Figure 4**-Mapping protein sequences to bag representation for MIL for solving problems in case studies. In all three problems, the actual minimal residue area responsible for the characteristic of interest lies inside the wider annotated area. However, it is not known in prior which part inside the annotated region is responsible for the property. Hence, we take smaller overlapping windows as examples from the annotated region to form a positive bag. Windows from rest of the sequence are used to form negative bags.

## 4.1 Binding Site Prediction in Calmodulin Binding Proteins

Calmodulin (CaM) is a calcium binding protein that plays a significant role in many vital cellular functions like metabolism, immune response, muscle contractions etc. in Eukaryotes (Chafouleas et al. 1982), (Walsh 1983). The fact that it is a highly conserved protein across all the eukaryotes adds to its significance. Identification of proteins that bind CaM and their CaM binding sites can help biologists understand the functions in which CaM is involved at a molecular level. Accurate identification of CaM binding site in a protein in the wet-lab is expensive and labor-intensive. Therefore, we have developed a machine learning based solution for prediction of CaM binding sites called CaMELS (CaM intEraction Learning System) (Abbasi et al. 2017).

CaMELS uses pyLEMMINGS based multiple instance learning for classification of CaM-binding and non-binding sequence windows. For this purpose, our training dataset was taken from the CaM Target Database (Yap et al. 2000) and contains 153 non-redundant CaM binding proteins with annotated binding site regions. However, due to limitations of wet-lab assays for identification of protein interactions, these annotations cover a region much larger than the minimal set of residues that are truly involved in the interaction. We take sequence windows of length 21 as examples and model the annotated binding site region of a protein as a positive bag. Windows from rest of the sequence are taken as negative instances and are grouped into negatively labeled bags. A number of features representations of the sequence windows have been used. These include: Amino Acid compositions (AAC) (Saidi, Maddouri, and Mephu Nguifo 2010), Position Dependent 1-Spectrum features (PD-1) (Minhas and Ben-Hur 2012), Position Dependent BLOSUM-62 features (Eddy and others 2004) and Propy features (Cao, Xu, and Liang 2013), etc. Further details of the experiment are presented in (Abbasi et al. 2017).

We have used leave one protein out cross validation for assessing the accuracy of different classification schemes including pyLEMMINGS. In this approach, a machine learning model is trained using all but one protein which is used for testing and this process is repeated for all proteins one by one. The performance metrics used for the evaluation of the model include Area Under Receiver Operating Characteristic Curve (AUC-ROC) (Fawcett 2006) and AUC-ROC$_{0.1}$. Area under the ROC curve is a measure of the probability that a randomly chosen positive example will be ranked higher than a randomly chosen negative example. AUC-ROC$_{0.1}$ is the area under the ROC curve up to the first 10% false positives. It tells about the sensitivity of the classifier at high precision.

## 4.2 Prediction of Prion Proteins

We have used pyLEMMINGS for the prediction of prion proteins and their prion-forming domains in proteins in our recently published method pRANK (Minhas, Ross, and Ben-Hur 2017). Prions are proteins that provide an alternate mode of genetic transfer without any involvement of nucleic acids (Prusiner 2012). Due to their unusual biological characteristics and their pathological significance, the computational prediction of prionogenic proteins and their prion forming domains is a very important area of research.

pRANK uses the dataset generated by Alberti et al. (Alberti et al. 2009) which contains a number of prion proteins for which prion domains have been localized. However, the exact fragments forming prions are not known for these proteins. The data set also contains a set of non-prion forming proteins. Therefore, the problem has been modeled using multiple instance learning. For a given protein, sequence windows of size 41 are taken as examples such that windows within an annotated prion forming region are taken as a positive bag and the remaining ones constitute a negative bag. Amino Acid Composition (AAC) (Saidi, Maddouri, and Mephu Nguifo 2010) features are computed for each window. A ranking style multiple instance classifier was trained using this data set and used for both prion classification and domain localization. The maximum scoring window in a test protein is used as the location of the prion domain whereas its score indicates the predicted propensity of prion formation. Evaluation was performed using the Leave One Protein Out cross validation model. AUC-ROC and AUC-PR were used as the evaluation metrics for comparison with classical classification schemes and other existing methods such as PAPA (Ross et al. 2013), PLAAC (Lancaster et al. 2014) and PrionW (Zambrano et al. 2015).

We evaluated the methods on yeast proteome dataset also and compared our methods with PAPA (Ross et al. 2013), PrionW (Zambrano et al. 2015), Michelitsh-Weissman (MW) score (Minhas, Ross, and Ben-Hur 2017) and the PLAAC-Log Likelihood Ratio (LLR) (Lancaster et al. 2014). In this problem, we defined prion positive proteins as positive bags and the rest as negative bags. For this experiment, bootstrapping was used for evaluation. i.e., one prion positive protein is held out and we train the classifier on a randomly selected subset of negative examples. The held-out protein and the rest of the negative proteins, that were not used in training, are then used in evaluation. This process was repeated for all positive bags. The whole procedure was repeated several times and the average ROC was reported. AUC-ROC and AUC-PR were used as the evaluation metrics. Further details of the experiment can be found in (Minhas, Ross, and Ben-Hur 2017).

## 4.3 Prediction of Amyloid Cores

Amyloids are collections of proteins folded into structures that cause many copies of the proteins to clump together and form fibrils (Gebbink et al. 2005). Formation of amyloids is associated with many neurodegenerative diseases such as Alzheimer's, Parkinson's, Huntington's disease, etc (Gebbink et al. 2005). We have used pyLEMMINGS for *in silico* prediction of amyloid cores in protein sequences (Munir and Minhas, n.d.). The study focuses on three amyloid-formation related problems, each of which is discussed in the following sections.

### 4.3.1 Prediction of Amyloid Proteins

In this task, the goal is to develop a system that predicts whether a given protein sequence can form amyloids or not. For this purpose, the training and evaluation dataset has been taken from Família et al. (Família et al. 2015). The dataset consists of two parts. The first part (DS-1) contains 304 hexapeptide sequences of which 168 have been experimentally tested to form amyloids and are taken as positive whereas the remaining are negative for amyloid formation. The other part of this dataset (DS-2) consists of 483 full-length protein sequences labeled positive (341 sequences) or negative for amyloid formation (142 sequences). The exact regions responsible for amyloid formation have not been annotated in DS-2. This gives rise to labeling ambiguities that can be modeled using multiple instance learning. To be able to use this data together with DS-1 in our predictor, we model each protein as a bag with an associated label by taking sequence windows of length six in a protein as instances. Each instance is represented by a 20-dimensional feature vector consisting of amino acid composition of the sequence. The propensity of a protein to form amyloids is taken to be the prediction score of the maximum scoring hexamer in it and the location of the window is predicted as part of the amyloid core. To provide a fair comparison of our method with existing amyloid prediction techniques which typically perform evaluation on dataset 2 only, we used 5-Fold cross validation over DS-2 for evaluation of our model. DS-1 was made a part of training data in all the folds and was not used in testing. Existing techniques used in our comparison include simple SVM, MetAmyl (Emily, Talvas, and Delamarche 2013), AGGRESCAN (Conchillo-Solé et al. 2007), APNN (Família et al. 2015) and Waltz (Oliveberg 2010). We used AUC-ROC as the performance metric for comparison.

### 4.3.2 Prediction of Amyloid Hotspots

Amyloid hotspots are subsequences in a protein that are responsible for amyloid formation. In this task, our goal was to assess the accuracy of our amyloid prediction model for identifying hotspots in a protein sequence. An additional dataset (DS-3) was used in this task (Emily, Talvas, and Delamarche 2013). This dataset comprises of 33 protein sequences with annotated hotspots. We trained our classifier using DS-1 and DS-2 and tested it over DS-3. The example with the highest score in a test bag was taken as the predicted hotspot. We compared our methods with existing techniques including simple SVM, MetAmyl (Emily, Talvas, and Delamarche 2013), AGGRESCAN (Conchillo-Solé et al. 2007) and APNN (Família et al. 2015). Here again, AUC-ROC was used as the performance metric for comparison.

### 4.3.3 Prediction of Effects of Mutations on Amyloid Formation Propensity

Mutations in a sequence can affect the propensity of a protein for amyloid formation (Merlini and Bellotti 2003). Our goal in this study was to predict the effects of mutation on propensity of amyloid formation for a protein. We used the dataset developed by Conchillo-Solé et al. (Conchillo-Solé et al. 2007) (henceforth called, DS-4) in addition to DS-1 and DS-2. DS-4 contains both wild-type and mutated versions of a number of proteins together with the effect (increase or decrease in amyloid formation) of those mutations. We modeled this prediction problem using multiple instance ranking. The proposed model was evaluated using Leave One Protein Out cross validation over DS-4 such that one protein from this data set is held out for testing whereas remaining proteins together with DS-1 and DS-2 are used for training.

## 5 Results and Discussion

In this section, we present and discuss the results over the benchmark datasets and the three protein labeling problems described in the previous section.

### 5.1 Results on Benchmark Datasets

The accuracy and AUC-ROC scores obtained from 10-fold cross validation over the two benchmark datasets are presented in Table 2. Five cross-validation runs of all the algorithms were performed. The average performance scores with their standard deviations for the five runs have been presented in the table. We compare the accuracy of our methods with other large-margin multiple instance learning methods including MILES (Chen, Bi, and Wang 2006), mi-SVM (Andrews, Tsochantaridis, and Hofmann 2003), MI-SVM (Andrews, Tsochantaridis, and Hofmann 2003), DD-SVM (Chen and Wang 2004), MissSVM (Zhou and Xu 2007), AW-SVM (Gehler and Chapelle 2007) and AL-SVM (Gehler and Chapelle 2007). The results for MILES, mi-SVM, MI-SVM and DD-SVM have been taken from (Chen, Bi, and Wang 2006). The rest have been taken from (Nguyen et al. 2013). AUC-ROC scores for other methods were not available. As shown in Table 2, our locally linear methods outperform other methods in terms of accuracy. Even simple linear methods also give accuracies comparable to the best existing large margin methods.

Apart from producing better accuracy than other large margin methods, pyLEMMINGS gives significantly better running times in comparison to heuristic based methods mi-SVM and MI-SVM for the larger dataset MUSK-2. The comparison is shown in Figure 5. The running time for MILES has been taken from (Chen, Bi, and Wang 2006). Our linear methods run much faster than MILES without a loss in accuracy. The locally linear algorithms show a comparable running time with a much better accuracy in comparison to MILES.

## 5.2 Results for Case Studies

For all the three problems discussed in section 2.6, pyLEMMINGS has proven to outperform the state of the art methods. In this section, we present the results for each of the problems.

**Table 2-** Percentage Accuracy and AUC-ROC scores of different methods over MUSK-1 and MUSK-2. Results for methods in pyLEMMINGS were taken by averaging the scores over five 10-Fold Cross Validation runs. Standard Deviations over the five runs are also mentioned.

| | Methods | Accuracy (%) | | AUC-ROC (%) | |
|---|---|---|---|---|---|
| | | **MUSK-1** | **MUSK-2** | **MUSK-1** | **MUSK-2** |
| pyLEMMINGS | Linear Classification | 87.25 ± 0.50 | 88.46 ± 1.49 | 85.38 ± 0.50 | 89.56 ± 1.37 |
| | Linear Ranking | 87.75 ± 1.46 | 90.00 ± 0.70 | 86.25 ± 1.94 | 89.53 ± 0.42 |
| | Locally Linear Classification | **92.25 ± 1.22** | 91.10 ± 1.80 | 90.13 ± 1.50 | 89.60 ± 2.17 |
| | Locally Linear Ranking | 91.25 ± 1.37 | **93.32 ± 1.24** | **91.88 ± 1.31** | **93.15 ± 0.44** |
| MILES (Chen, Bi, and Wang 2006) | | 86.3 | 87.7 | - | - |
| mi-SVM (Andrews, Tsochantaridis, and Hofmann 2003) (Chen, Bi, and Wang 2006) | | 87.4 | 83.6 | - | - |
| MI-SVM (Andrews, Tsochantaridis, and Hofmann 2003) (Chen, Bi, and Wang 2006) | | 77.9 | 84.3 | - | - |
| AW-SVM (Gehler and Chapelle 2007) (Nguyen et al. 2013) | | 86.0 | 84.0 | - | - |
| AL-SVM (Gehler and Chapelle 2007) (Nguyen et al. 2013) | | 86.0 | 83.0 | - | - |
| MissSVM (Zhou and Xu 2007) (Nguyen et al. 2013) | | 87.6 | 80.0 | - | - |
| DD-SVM (Chen and Wang 2004) (Chen, Bi, and Wang 2006) | | 85.8 | 91.3 | - | - |

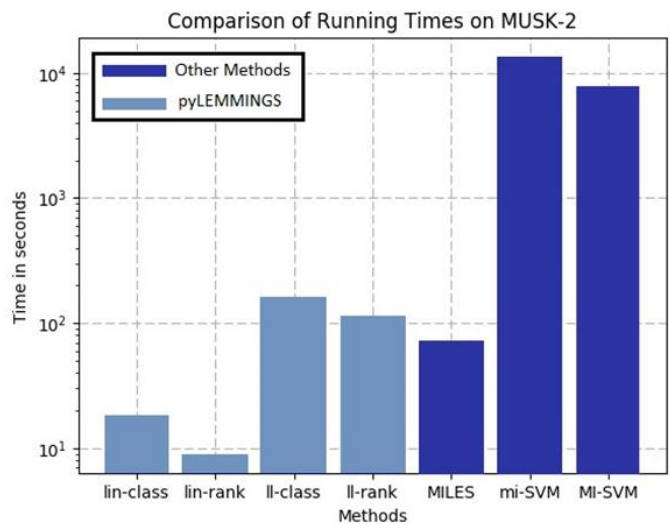

**Figure 5-** Comparison of running times of different methods on MUSK-2. lin-class, lin-rank, ll-class and ll-rank refer to the linear classifier, linear ranking, locally linear classifier and locally linear ranking algorithms in pyLEMMINGS respectively. MILES shows a comparable performance in terms of running time but out algorithms show much better classification accuracy.

**Table 3-** Percentage AUC-ROC and AUC-ROC 0.1 scores for different methods. pyLEMMINGS based method (CaMELS) shows better performance than the previous state of the art methods.

| Method | Features | AUC-ROC (%) | AUC-ROC$_{0.1}$ (%) |
|---|---|---|---|
| CaMELS using pyLEMMINGS (Linear Ranking) (Abbasi et al. 2017) | PD-BLOSUM | **99.2** | **79.0** |
| | AAC+PD-1 | **98.9** | 77.6 |
| | PD-GT | 99.04 | 78.0 |
| | PD-1 | 98.4 | 76.2 |
| | Propy | 98.0 | 74.7 |
| | AAC | 97.9 | 72.3 |
| MI-1 (Minhas and Ben-Hur 2012) | AAC+PD-1 | 96.9 | 59.0 |
| mi-SVM (Andrews, Tsochantaridis, and Hofmann 2003) | AAC+PD-1 | 96.2 | 55.6 |
| SVM | AAC+PD-1 | 95.9 | 55.1 |

**Table 4-** AUC-ROC and AUC-PR Results for Prion Activity Prediction on Alberti dataset

| Method | AUC-ROC (%) | AUC-PR (%) |
|---|---|---|
| pRANK using pyLEMMINGS (Linear Ranking) | **96.8** | **96.8** |
| mi-SVM (Andrews, Tsochantaridis, and Hofmann 2003) | 92.2 | 90.4 |
| SVM | 87.4 | 87.8 |
| Random Forests | 88.0 | 90.6 |
| PAPA (Ross et al. 2013) | 95.1 | 96.8 |
| PrionW (Zambrano et al. 2015) | 86.7 | 89.8 |
| PLAAC (Lancaster et al. 2014) | 68.7 | 74.7 |

**Table 5-** AUC-ROC and AUC-PR Results for Prion Activity Prediction on yeast proteome dataset

| Method | AUC-ROC (%) | AUC-PR (%) |
|---|---|---|
| pRANK using pyLEMMINGS (Linear Ranking) | **99.2** | **53.3** |
| MW | 99.2 | 22.4 |
| PLAAC-LLR (Lancaster et al. 2014) | 99.2 | 27.0 |
| PAPA (Ross et al. 2013) | 98.4 | 24.6 |
| PrionW (Zambrano et al. 2015) | 98.4 | 17.5 |

### 5.2.1 Binding Site Prediction in Calmodulin Binding Proteins

CaMELS (Abbasi et al. 2017) using pyLEMMINGS for binding site prediction of Calmodulin binding proteins has shown to outperform the previously published methods both in terms of AUC-ROC and AUC-ROC$_{0.1}$ as shown in Table 3. CaMELS, has outperformed MI-1 (Minhas and Ben-Hur 2012), mi-SVM (Andrews, Tsochantaridis, and Hofmann 2003) and SVM by producing an AUC-ROC of 98.9% when Amino Acid Composition and Position Dependent 1-Spectrum features were used.

Using a different feature set of Position Dependent BLOSUM-62 features, further improvement was recorded. The AUC-ROC increased to 99.2%. To analyze the True-Positive Rate over small False-Positive Rate AUC-ROC$_{0.1}$ have also been computed. Previously, the best AUC-ROC$_{0.1}$ over Amino Acid Composition and Position Dependent 1-Spectrum features was reported to be 59.0% by MI-1. CaMELS over the same features has produced a much-improved AUC-ROC$_{0.1}$ of 77.6%. The AUC-ROC$_{0.1}$ was further improved to 79.0% when we used Position Dependent BLOSUM-62 features.

Overall, the pyLEMMINGS based method has shown a great improvement in prediction of binding sites in Calmodulin binding proteins.

### 5.2.2 Prediction of Prion Forming Proteins

Our pyLEMMINGS based predictor of prion forming protein, pRANK (Minhas, Ross, and Ben-Hur 2017), outperforms previous state of the art method. Our method produces an AUC-ROC and AUC-PR scores of 96.8% as shown in Table 4 using only Amino Acid Composition Features. It is interesting to note that both multiple instance learning based methods (mi-SVM and pRANKS) outperform classical machine learning techniques (SVM and Random Forests). pRANK performs at par with state of the art prion predictor on this dataset. However, pRANK offers significant improvement in prediction performance when predicting prions in the yeast proteome as shown in Table 5.

### 5.2.3 Prediction of Amyloids

Results of the three experiments for prediction of amyloids using pyLEMMINGS are given in Table 6. Simple linear classifier implemented in pyLEMMINGS, when used for prediction of amyloid proteins produced an AUC-ROC 88.1%. Here too, multiple instance learning results in improvement of prediction performance in comparison to a conventional classifier. The result is comparable to the AUC-ROC achieved by the state of the art, much more complex method, MetAmyl (Emily, Talvas, and Delamarche 2013) that gave an AUC-ROC of 88.3%.

For the task of hotspot prediction, the pyLEMMINGS based linear classification method outperforms existing amyloid hotspot prediction methods (Família et al. 2015) by producing an AUC-ROC of 98.0%. The AUC-ROC score reported for the previous state of the art method APPNN (Família et al. 2015) is 97.3%.

Prediction of mutation effects causing Amyloid formation was performed using linear ranking in pyLEMMINGS. In comparison to the state-of-the-art method AGGRESCAN (Conchillo-Solé et al. 2007), whose AUC-ROC was 94.8%, our algorithm produced an AUC-ROC score of 97.0%.

Overall, pyLEMMINGS based algorithms have shown to produce better results using much simpler models for prediction of Amyloid cores as compared to the other more complex models.

**Table 6-** AUC-ROC scores for Amyloid Classification, Hotspot prediction and effects of mutations on propensity of amyloid formation. pyLEMMINGS based methods outperform the state of the art methods for hotspot prediction and effects of mutation prediction. Our linear classifier produces comparable AUC-ROC to the state of the art technique

| Task | Method | AUC-ROC (%) |
|---|---|---|
| Amyloid Classification | pyLEMMINGS- Linear Classification | 88.1 |
| | pyLEMMINGS- Linear Ranking | 85.9 |
| | Linear SVM | 83.1 |
| | MetAmyl (Emily, Talvas, and Delamarche 2013) | **88.3** |
| | AGGRESCAN (Conchillo-Solé et al. 2007) | 79.5 |
| | APPNN (Família et al. 2015) | 87.9 |
| | Waltz (Oliveberg 2010) | 71.3 |
| Amyloid Hotspot Prediction | pyLEMMINGS- Linear Classification | **98.0** |
| | pyLEMMINGS- Linear Ranking | 97.8 |
| | Linear SVM | 96.8 |
| | MetAmyl (Emily, Talvas, and Delamarche 2013) | 96.8 |
| | AGGRESCAN (Conchillo-Solé et al. 2007) | 94.1 |
| | APPNN (Família et al. 2015) | 97.3 |
| Effects of Mutation Prediction | pyLEMMINGS- Linear Classification | 82.4 |
| | pyLEMMINGS- Linear Ranking | **97.0** |
| | AGGRESCAN (Conchillo-Solé et al. 2007) | 94.8 |

# 6 Conclusion

We have presented stochastic sub-gradient optimization solvers for linear and locally-linear multiple instance learning. The learning problems discussed include classification and ranking. Apart from giving a direct solution to the respective large margin optimization problems, our approaches result in improved accuracy and a many-fold improvement in run-times in comparison to heuristic solvers.

As an alternative to conventional kernelization, we have proposed novel locally linear encoding based algorithms for multiple instance learning. Also, we have developed an open source software suite implementing all the discussed techniques. The pyLEMMINGS software provides Python implementations of the proposed algorithms. The software has been used to solve three real world problems. pyLEMMINGS in all three case studies has shown to produce better results than the state of the art algorithms.